\UseRawInputEncoding{}
\documentclass[conference]{IEEEtran}
\usepackage{cite}
\usepackage{amsmath,amssymb,amsfonts, amsthm}
\usepackage{algorithm}
\usepackage{algorithmic}
\usepackage[final]{graphicx}
\usepackage{booktabs}   
\usepackage{textcomp}
\usepackage{xcolor}
\usepackage{url}
\def\BibTeX{{\rm B\kern-.05em{\sc i\kern-.025em b}\kern-.08em
    T\kern-.1667em\lower.7ex\hbox{E}\kern-.125emX}}

\begin{document}

\title{GenEx: A Graph-Based Representational Paradigm for SARS-CoV-2 Variant Detection via Codon Co-occurrence Networks\\}

\author{
\begin{tabular}{cc}
\begin{tabular}{@{}c@{}}
\textbf{Arefin Amin} \\
\textit{ECE Department} \\
\textit{North South University} \\
Dhaka, Bangladesh \\
arefin.amin@northsouth.edu\\
\end{tabular} &
\begin{tabular}{@{}c@{}}
\textbf{Labiba Faiza Karim} \\
\textit{ECE Department} \\
\textit{North South University} \\
Dhaka, Bangladesh \\
labiba.karim@northsouth.edu\\
\end{tabular}
\begin{tabular}{@{}c@{}}
\textbf{M. Monir Uddin} \\
\textit{Department of Math. \& Phy.} \\
\textit{North South University} \\
Dhaka, Bangladesh \\
monir.uddin@northsouth.edu\\
\end{tabular}
\end{tabular}
}
\maketitle

\begin{abstract}
Genomic analysis on viruses such as SARS-CoV-2 variants: Beta, Gamma, Delta, and Omicron is heavily dominated by classical bioinformatics methods, including Sequence Alignment, Phylogenetic Analysis, and Mutation Frequency Statistics. These approaches use pairwise codon or nucleotide distance matrices to analyze gene sequences, treating them as linear strings rather than capturing their complex contextual interdependencies. We proposed GenEx, a pipeline that converts raw gene sequences into codon co-occurrence graphs and extracts more than 25 graph features. Our two most prominent techniques for graph generation and feature extraction are MSCG (Multi-Scale Codon Co-occurrence Graph) and LAPCG (Linear-time Adjacency PMI Codon Graph). Using these algorithms, we treated codon sequences as structured symbolic vocabularies interpretable to codon co-occurrence graph analysis, a representational paradigm borrowed from computational linguistics. Another major contribution includes implementing a spectral graph feature extraction using Singular Value Decomposition (SVD), using the squared singular value ($\sigma^2$) instead of the traditionally used eigenvalue, which helped us to amplify the separation between dominant and subdominant spectral components, thereby enhancing inter-class separability in downstream classification.  And to further demonstrate that our method works, we trained 23 benchmarked ML models against the latest SARS-CoV-2 variants, achieving remarkable results in detecting all SARS-CoV-2 variants.
\end{abstract}

\begin{IEEEkeywords}
Sequence Alignment, Phylogenetic Analysis, Multi-Scale Codon Co-occurrence Graph, Computational linguistics, Spectral graph features.
\end{IEEEkeywords}

\section{Introduction}
\label{sec:intro}
Understanding viral genome evolution is a fundamental problem in computational biology, with direct implications for evolutionary studies and vaccine monitoring. While the SARS-CoV-2 pandemic has enabled the collection of large-scale big data on viral genome sequences, conventional methods: phylogenetic tree reconstruction, mutation frequency statistics, and multiple sequence alignment (MSA), heavily focus on retrospective evolutionary events, stationarity, site-independence, and a linear mutation process; thus, failing to capture the complex relationships and interactions within genomic sequences. To resolve these bottlenecks, we propose shifting from the static linear analysis to structural graph models that capture the topological structure of genetic sequences. By constructing codon co-occurrence graphs, where nodes and edges represent codons and their local relationships, respectively, we can capture neighborhood patterns and mutation tendencies that are invisible to traditional approaches. 

We propose GenEx, a novel Graph-Based Representation framework that models viral genomes as codon-level co-occurrence graphs and identifies distinct topological signatures and spectral properties associated with different viral variants. GenEx extracts spectral and topological features by computing direct-neighbor interactions, generalizing these associations across multiple co-occurrence scales and positional segment encodings; a data-driven graph paradigm for clade classification that prioritizes mathematical explainability and computational efficiency over black-box modeling. This architecture treats variant detection as a measurable task within graph theory, providing a robust, data-driven methodology for classifying viral evolution. We initially focused on SARS-CoV-2 because large-scale genomic sequences were available. This proposed framework will be designed to be adaptable, thus it can extend to other viral families and higher organisms.

\subsection{Motivation:}
The motivation for this study originated from the following reasons:
\begin{enumerate}
    \item \textbf{Limitations of existing genome analysis methods:}
    Traditional gene analysis methods, such as phylogenetic and statistical mutation models, are often constrained by assumptions of linear, site-independent mutation patterns and fail to capture the complex contextual interdependencies within the genomic sequences.
    \item \textbf{Need for ``Alignment-Free'' Surveillance:}
    With the exponentially growing viral data, alignment-free frameworks are needed to characterize variants through their topological signatures, which enables rapid identification of emerging Variants of Concern (VOC) without bottlenecks.
    \item \textbf{Preparation for Future Pathogens: }
    The research provides a mathematical foundation for the evolution of future pathogens, pandemics, and vaccines by developing a generalizable, alignment-free model applicable to pathogens and higher organisms.
\end{enumerate}

\subsection{Contributions:}
Our work concentrates on the following aspects:
\begin{enumerate}
    \item \textbf{Dual-Paradigm Graph Construction Algorithms:}
    We proposed two novel algorithms, LAPCG and MSCG, to extract codon interaction networks from linear gene sequences with $\mathcal{O}(n)$ and $\mathcal{O}(n^2)$ complexity, respectively, while maintaining a 98.75\% accuracy.
    
    \item \textbf{Spectral Feature Engineering and Mathematical Framework:}
    We introduced a Singular Value Decomposition (SVD) based framework that derives seven distinct spectral features from squared singular values ($\sigma^2$), providing a numerically stable representation of genomic interaction systems with high inter-class separability.

    \item \textbf{Mathematical Explainability and Biological Interpretation:}
    We were able to provide a rigorous interpretation of viral clades by correlating their biological interpretations with numerical graph properties; thus, we can view the structural mechanisms of viral evolution and variant-specific interaction dynamics.
\end{enumerate}

\section{Related Works}

Recent progress in computational biology has made genome analysis faster and more scalable. This section briefly reviews the work streams most relevant to GenEx.

\begin{itemize}

\item \textbf{Sequence-based Deep Learning Models.} CNN- and Transformer-based models treat genomes as linear nucleotide strings for regulatory and mutational prediction \cite{zhou2015predicting, avsec2021effective}. They are effective for sequence pattern recognition, but usually ignore codon-level structure and rely on post-hoc explanation.

\item \textbf{Genome Graphs and Pangenome Representations.} Genome and pangenome graphs encode population variation with sequence-segment nodes and variant-path edges \cite{garrison2018variation}. These frameworks are useful for representation, but are often non-learning, mostly static, and not built for codon-level mutation feature learning.

\item \textbf{Graph Neural Networks in Bioinformatics.} GNNs perform well on molecular graphs, protein--protein interaction networks, and gene regulation tasks \cite{zitnik2018modeling, gilmer2017neural}. However, most studies are static and rarely model codon-level evolutionary dynamics directly.

\item \textbf{Evolutionary and Phylogenetic Models.} Classical tools such as PAM \cite{dayhoff1978atlas} and BLOSUM \cite{henikoff1992amino} are biologically interpretable and foundational. Their limitation is dependence on assumptions such as stationarity and site independence, which can miss nonlinear mutation behavior.

\item \textbf{Explainable Genomic AI.} Attribution methods (e.g., DeepLIFT \cite{shrikumar2017learning}) and attention analysis are widely used for interpretability. In practice, most are sequence-centric and post-hoc, giving limited structural insight into codon relationships.

\item \textbf{Ab initio and Pipeline-based Gene Annotation.} Systems including GENSCAN \cite{burge1997prediction}, GeneMark \cite{lomsadze2005gene}, AUGUSTUS \cite{stanke2008using}, GlimmerHMM \cite{majoros2004tigrscan}, BRAKER2 \cite{hoff2019whole}, MAKER2 \cite{holt2011maker}, and Helixer \cite{holst2023helixer} are strong annotation tools. Their objective, however, is gene-structure annotation, not variant classification through codon co-occurrence topology.

\item \textbf{Our Approach (GenEx).} GenEx models each genome as a codon co-occurrence graph, where nodes are codons and edges encode positional/mutational coupling. We then classify variants using graph-derived features, yielding interpretable structure-aware signals that can be extended to amino-acid and structure-level graphs.

\end{itemize}

\begin{table}[t]
\centering
\caption{Comparison of Prior Work and Proposed Method}
\label{tab:prior_comparison}
\resizebox{\columnwidth}{!}{%
\begin{tabular}{lll}
\toprule
\textbf{Aspect} & \textbf{Prior Work} & \textbf{Our Method} \\
\midrule
Biological unit      & Nucleotide / segment       & Codon \\
Representation       & Sequence / static graph    & Co-occurrence graph \\
Temporal modeling    & No                         & Yes \\
Learning framework   & CNN / Transformer / rules  & ML on graph features \\
Explainability       & Post-hoc                   & Intrinsic (graph-based) \\
Multi-task reuse     & Limited                    & Yes \\
\bottomrule
\end{tabular}
}
\end{table}

\section{Dataset}
\label{sec:data}
We used the following publicly available SARS-CoV-2 genome sequence datasets from the National Center for Biotechnology Information (NCBI)\cite{ncbi} Virus database to train, validate, and test the GenEx framework:

\subsection{Dataset Collection Process:}
The SARS-CoV-2 genome sequence data for this research were retrieved using the ncbi-datasets command-line interface. We used the following fetch commands to fetch the variants in ZIP format:

\begin{itemize}
    \item \textbf{Beta:} {\raggedright\texttt{datasets download virus genome taxon SARS-CoV-2 --lineage B.1.351}\cite{ncbi}\par}
    \item \textbf{Delta:} {\raggedright\texttt{datasets download virus genome taxon SARS-CoV-2 --lineage B.1.617.2}\cite{ncbi}\par}
    \item \textbf{Gamma:} {\raggedright\texttt{datasets download virus genome taxon SARS-CoV-2 --lineage P.1}\cite{ncbi}\par}
    \item \textbf{Omicron:} {\raggedright\texttt{datasets download virus genome taxon SARS-CoV-2 --lineage B.1.1.529}\cite{ncbi}\par}
\end{itemize}

{\raggedright After fetching the ZIP files, we extracted \path{beta_sequences.csv}, \path{delta_sequences.csv}, \path{gamma_sequences.csv}, and \path{omicron_sequences.csv}. From each CSV file, we selected 1250 randomized sequences, yielding 5000 viral genome sequences in total.\par}

\subsection{Dataset Characteristics and Structure:}
Each raw CSV file and the extracted sequence CSV files (1250 sequences per variant) had 3 columns: header, sequence, and sequence\_length. 
\begin{itemize}
    \item \textbf{header}: Contained the NCBI accession ID for the respective genome sequence, the symptoms, virus name (SARS-CoV-2), Host/Carrier, etc., as a single string. 
    \item \textbf{sequence}: Contained gene sequences.
    \item \textbf{sequence\_length}: Contained an integer representing the length of each gene sequence.
\end{itemize}

\subsection{Data Pre-processing}
After downloading the data, we began our pre-processing -

\begin{itemize}
    \item First we took first 10 character deicarding the rest from the header column of the datasets as those first 10 characters made up the accession ID.
    \item The Gnomes are officially stored as cDNA (A,T,G,C) sequences even though SARS-CoV-2 is an RNA (A,T,U,C) virus. Thus, to ensure compatibility with the main stream we avoided unnecessary conversion and maintained the cDNA sequences.
    \item Next, we capitalized the Gene sequences and removed unnecessary sequences and characters that are not relevant to our current work. 
    \item We found 20 thousand clean sequences with unique accession ID from each of these variants. 
\end{itemize}

\section{Methodology}

\subsection{Overview}
GenEx converts each SARS-CoV-2 nucleotide sequence into a codon graph, extracts structural and spectral descriptors, and then uses these descriptors for variant-level statistical analysis and machine learning classification. We used two graph construction strategies: \textbf{LAPCG} for fast local codon interactions and \textbf{MSCG} for multi-scale codon context modeling.

\subsection{End-to-end Pipeline}
\begin{enumerate}
    \item Convert each sequence into codons using fixed-frame triplets.
    \item Build a weighted codon graph using either LAPCG or MSCG.
    \item Compute graph features (topological, path-based, centrality, and spectral).
    \item Run significance tests (ANOVA) across variants.
    \item Benchmark ML models on extracted feature vectors.
\end{enumerate}

\begin{figure*}[!t]
\centering
\IfFileExists{images/struct_col.png}{%
  \includegraphics[width=0.92\textwidth]{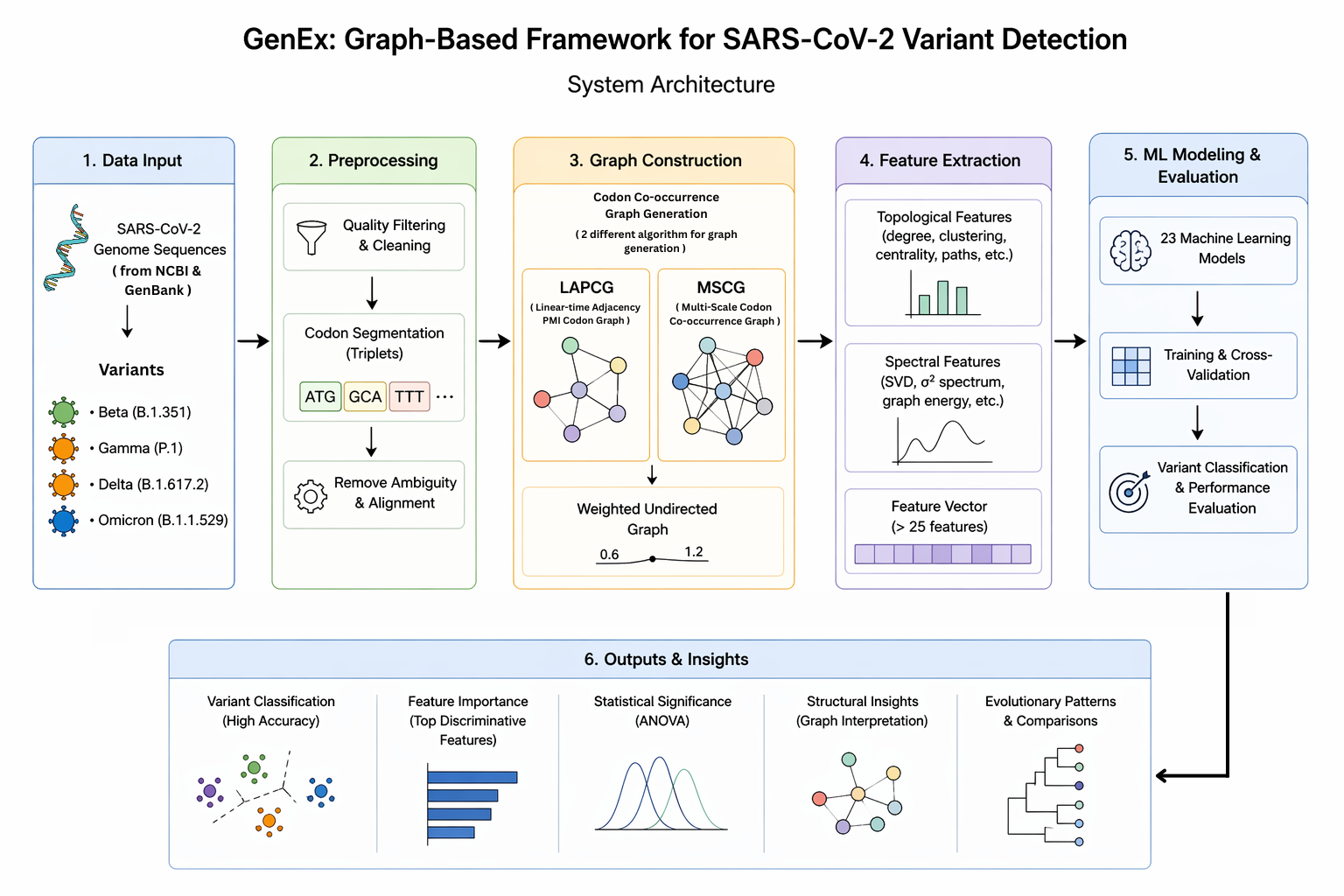}%
}{%
  \fbox{\parbox{0.95\linewidth}{\centering\vspace{1cm}Missing: images/struct\_col.png\vspace{1cm}}}%
}
\caption{Overview of the GenEx framework for SARS-CoV-2 variant detection using a graph-based approach. The pipeline begins with genome sequence acquisition from NCBI GenBank, followed by preprocessing steps including quality filtering, codon segmentation into triplets, and sequence alignment. Codon co-occurrence graphs are then constructed using two methods---LAPCG (linear-time adjacency-based) and MSCG (multi-scale co-occurrence)---to form weighted undirected graphs. From these graphs, both topological and spectral features are extracted to create high-dimensional feature vectors. These features are used to train and evaluate multiple machine learning models for variant classification. The framework outputs include high-accuracy variant predictions, feature importance analysis, statistical validation, structural graph insights, and evolutionary pattern comparisons.}
\label{fig:architecture}
\end{figure*}

\subsection{Codon Segmentation}
Given a nucleotide sequence
\[
S = (n_1, n_2, \dots, n_L),
\]
we form codons as non-overlapping triplets:
\[
C = (c_1, c_2, \dots, c_{\lfloor L/3 \rfloor}), \quad
c_i = (n_{3i-2}, n_{3i-1}, n_{3i}).
\]
Ambiguous symbols (\textit{e.g.}, \texttt{N}) are removed before graph construction.

\subsection{LAPCG: Linear-time Adjacency PMI Codon Graph}
LAPCG models only adjacent codon pairs $(c_i, c_{i+1})$, which provides linear complexity in sequence length and works well for large-scale data.

\begin{algorithm}[t]
\caption{LAPCG Graph Construction}
\label{alg:lapcg}
\begin{algorithmic}[1]
\REQUIRE Codon sequence $C=(c_1,\dots,c_N)$
\ENSURE Undirected weighted graph $G=(V,E,W)$
\STATE Initialize node-frequency map $F$ and adjacent-pair map $M$
\FOR{$i=1$ to $N$}
    \STATE $F(c_i) \leftarrow F(c_i)+1$
\ENDFOR
\FOR{$i=1$ to $N-1$}
    \STATE $u \leftarrow c_i,\; v \leftarrow c_{i+1}$
    \STATE $M(u,v) \leftarrow M(u,v)+1$ \COMMENT{unordered pair}
\ENDFOR
\STATE $V \leftarrow$ unique codons in $C$
\FORALL{pairs $(u,v)$ with $M(u,v)>0$}
    \STATE $p(u,v)\leftarrow \frac{M(u,v)}{N-1}$,\;\;
           $p(u)\leftarrow \frac{F(u)}{N}$,\;\;
           $p(v)\leftarrow \frac{F(v)}{N}$
    \STATE $\mathrm{PMI}(u,v)\leftarrow \log\!\left(\frac{p(u,v)}{p(u)p(v)}\right)$
    \IF{$\mathrm{PMI}(u,v)>0$}
        \STATE Add edge $(u,v)$ with weight $\mathrm{PMI}(u,v)$
    \ENDIF
\ENDFOR
\RETURN $G=(V,E,W)$
\end{algorithmic}
\end{algorithm}

\subsection{MSCG: Multi-Scale Codon Co-occurrence Graph}
MSCG extends local adjacency by capturing co-occurrence across multiple codon distances. In our setup, we used scales $s \in \{1,2,3\}$ with decay weights $(1.0,0.5,0.25)$ and normalized PMI (NPMI).

\begin{algorithm}[t]
\caption{MSCG Graph Construction}
\label{alg:mscg}
\begin{algorithmic}[1]
\REQUIRE Codon sequence $C=(c_1,\dots,c_N)$, scales $\mathcal{S}=\{(s,\alpha_s)\}$
\ENSURE Multi-scale weighted graph $G=(V,E,W)$
\STATE Initialize node statistics and empty edge accumulator $A_e$
\STATE $V \leftarrow$ unique codons in $C$
\FORALL{$(s,\alpha_s)\in\mathcal{S}$}
    \FOR{$i=1$ to $N-s$}
        \STATE Observe pair $(c_i, c_{i+s})$ and update pair counts
    \ENDFOR
    \FORALL{observed pairs $(u,v)$}
        \STATE Compute $\mathrm{PMI}(u,v)$ and
        \[
        \mathrm{NPMI}(u,v)=\frac{\mathrm{PMI}(u,v)}{-\log p(u,v)}
        \]
        \IF{$\mathrm{NPMI}(u,v)>0$}
            \STATE $A_e(u,v)\leftarrow A_e(u,v)+\alpha_s \cdot \mathrm{NPMI}(u,v)$
        \ENDIF
    \ENDFOR
\ENDFOR
\FORALL{pairs $(u,v)$ in $A_e$}
    \STATE Add edge $(u,v)$ with accumulated weight $A_e(u,v)$
\ENDFOR
\RETURN $G=(V,E,W)$
\end{algorithmic}
\end{algorithm}

\subsection{Feature Extraction and Evaluation}
From each graph we extracted structural and spectral features, including number of nodes/edges, density, diameter, radius, average shortest path length, Wiener index, transitivity, clustering, centrality scores, graph energy, top singular-spectrum components, max-flow, and matching number. These were used for:
\begin{itemize}
    \item ANOVA-based inter-variant significance analysis,
    \item feature-importance analysis,
    \item model benchmarking across multiple ML families.
\end{itemize}

\section{Results}
\label{sec:results}

\subsection{Model Benchmarking}
Table~\ref{tab:benchmark_summary} summarizes the best-performing models for baseline PMI, LAPCG, and MSCG. MSCG achieved the strongest overall performance in our benchmark.

\begin{table}[t]
\centering
\caption{Benchmark summary: accuracy and runtime across graph construction methods.}
\label{tab:benchmark_summary}
\begin{tabular}{lccc}
\toprule
\textbf{Method} & \textbf{Best Model(s)} & \textbf{Accuracy} & \textbf{Runtime (s)} \\
\midrule
PMI baseline & CatBoost & 96.25\% & 272.3 \\
LAPCG & LightGBM, Grad.\ Boost & 96.25\% & 196.2 \\
MSCG & MLP, Bagging Classifier & \textbf{98.75\%} & \textbf{144.6} \\
\bottomrule
\end{tabular}
\end{table}

\subsection{Computational Efficiency of Graph Construction}

A critical practical advantage of our proposed methods is their computational efficiency relative to standard PMI-based graph construction. The PMI baseline, which computes global co-occurrence statistics over the entire sequence, requires \textbf{272.3\,s} to complete the full pipeline. LAPCG reduces this to \textbf{196.2\,s} (a \textbf{1.39$\times$ speedup}) by restricting edge computation to adjacent codon pairs, operating in strict $O(n)$ time. MSCG, despite modeling three co-occurrence scales simultaneously, achieves the lowest runtime of \textbf{144.6\,s}---a \textbf{1.88$\times$ speedup} over the PMI baseline---because the NPMI computation and accumulated edge weighting are implemented as a single-pass accumulation rather than a global normalization step. Crucially, this efficiency gain is not purchased at the cost of accuracy: MSCG also achieves the highest classification accuracy at 98.75\%, demonstrating that multi-scale structural modeling of codon co-occurrence is both faster and more discriminative than traditional PMI.

These runtime measurements were obtained on the full SARS-CoV-2 genome sequence dataset on a single CPU core, without parallelization. The MSCG pipeline---including graph construction, feature extraction, and ML classification---processes each genome in under 1\,s on average, making it suitable for real-time variant surveillance at scale.

\subsection{Comparison with Gene Detection and Variant Classification Methods}

Table~\ref{tab:comparison} situates GenEx within the broader landscape of genomic sequence analysis tools. We distinguish two categories: (i) traditional \textit{gene annotation} tools that predict gene structure from raw sequence (\textit{e.g.}, AUGUSTUS, GeneMark), and (ii) \textit{variant classification} methods that assign a class label to a given genome. Our GenEx falls into the second category. Accuracy figures for gene annotation tools are reported in terms of nucleotide-level sensitivity/specificity, while classification accuracy for variant detection methods corresponds to multi-class labeling performance.

\begin{table*}[t]
\centering
\caption{Comparison of GenEx with gene annotation tools and SARS-CoV-2 variant classification methods.
Gene annotation accuracies reflect nucleotide-level sensitivity; variant classification accuracies reflect multi-class accuracy.
$\dagger$~Whole-genome support assumes availability of a closely related reference.}
\label{tab:comparison}
\begin{tabular}{llllcc}
\toprule
\textbf{Method} & \textbf{Type} & \textbf{Task} & \textbf{Accuracy (reported)} & \textbf{Whole Genome} & \textbf{Reference} \\
\midrule
GENSCAN              & Ab initio HMM    & Gene annotation        & $\sim$70 - 80\% (protein-level)   & \checkmark & \cite{burge1997prediction} \\
GeneMark-ES/ET       & Ab initio (self-train) & Gene annotation   & 35.7 -- 75.8\% (nucleotide)        & \checkmark & \cite{lomsadze2005gene} \\
SNAP                 & Ab initio HMM    & Gene annotation        & 77 - 80\% (nucleotide)        & \checkmark & \cite{korf2004gene} \\
GlimmerHMM           & Ab initio HMM    & Gene annotation        & $\sim$9 -- 43\% (some datasets)   & \checkmark & \cite{majoros2004tigrscan} \\
MAKER2               & Annotation pipeline & Gene annotation     & 68.60\% (nucleotide)        & \checkmark & \cite{holt2011maker} \\
AUGUSTUS             & Ab initio + evidence & Gene annotation    & 82--92\% (gene-level)        & \checkmark & \cite{stanke2008using} \\
BRAKER2              & Hybrid (RNA-seq + HMM) & Gene annotation  & $>$AUGUSTUS ($+$2--3\%)      & \checkmark & \cite{hoff2019whole} \\
Helixer (DL)         & Deep learning    & Gene annotation        & 86.8\% (reported)          & \checkmark & \cite{holst2023helixer} \\
\midrule
$k$-mer + SVM        & Traditional ML   & Variant classification & $\sim$92.01\%               & \checkmark & Various \\
CNN-LSTM             & Deep learning    & Variant classification & $\sim$95--97\%               & Partial    & Various \\
\midrule
GenEx LAPCG (Ours)   & Graph + ML ($O(n)$) & Variant classification & 96.25\%                   & \checkmark & This work \\
\textbf{GenEx MSCG (Ours)} & \textbf{Graph + ML (multi-scale)} & \textbf{Variant classification} & \textbf{98.75\%} & \checkmark & This work \\
\bottomrule
\end{tabular}
\end{table*}

GenEx MSCG achieves the highest reported accuracy among all compared methods for its classification task, while also improving computational efficiency over standard PMI-based construction. It is important to note that gene annotation tools address a structurally different problem---predicting gene coordinates in an unannotated genome---and are therefore not directly comparable in terms of accuracy numbers. Nevertheless, placing GenEx alongside these tools provides useful context: our 98.75\% accuracy is achieved on a whole-genome, alignment-free basis without any reference sequence or transcriptomic data, which contrasts favorably with the reference-dependent or transcript-dependent nature of hybrid pipeline methods such as BRAKER2 and EVidenceModeler.

\subsection{Feature Importance and Statistical Significance}
MSCG emphasized both path-based and spectral properties. The highest ranked features were second eigenvalue, average shortest path length, Wiener index, top eigenvalue, and radius. ANOVA analysis further showed strong between-variant separation for radius ($F=69.74$, $p=3.14\times10^{-36}$), average shortest path length ($F=60.19$, $p=4.43\times10^{-32}$), top eigenvalue ($F=47.82$, $p=2.14\times10^{-26}$), and number of edges ($F=33.52$, $p=2.51\times10^{-19}$).

\subsection{Variant Structural Profiles}
\label{sec:variant_profiles}

\begin{figure*}[p] 
\centering
\IfFileExists{images/struct_col.png}{%
  \includegraphics[width=0.84\textwidth]{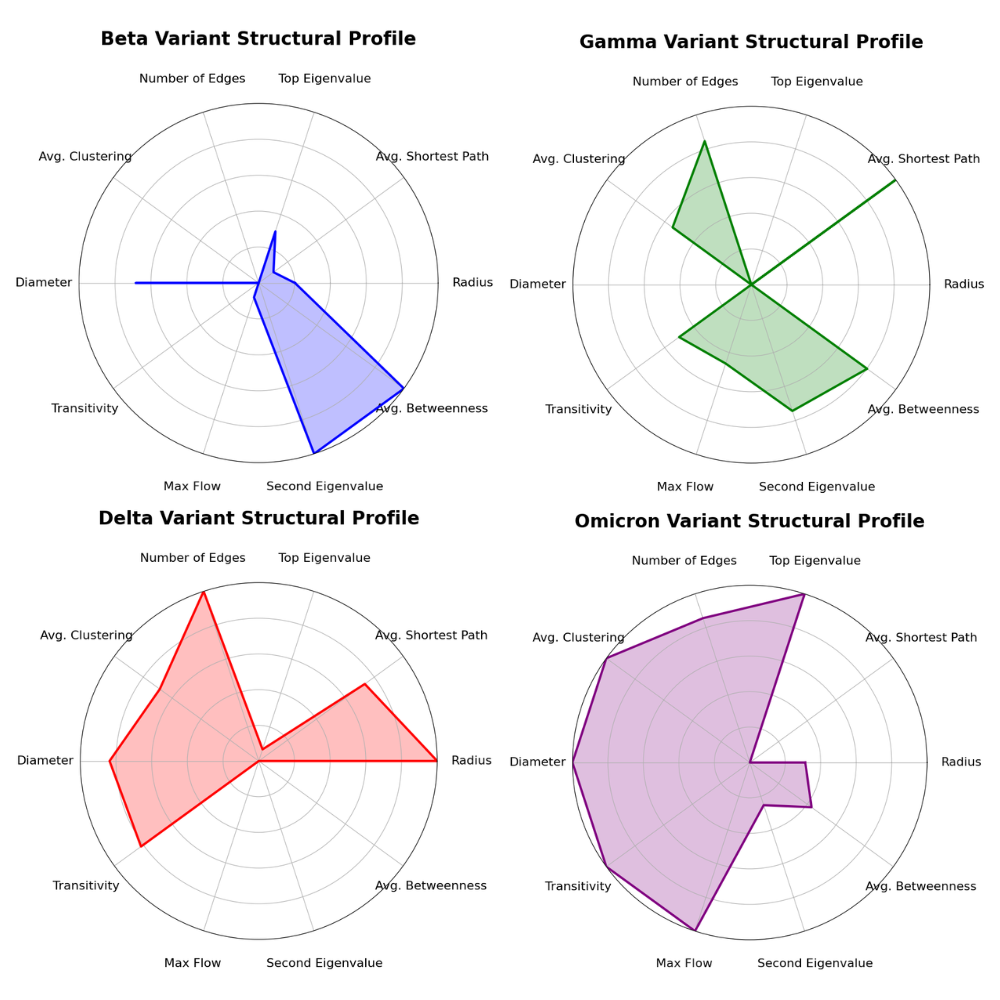}%
}{%
  \fbox{\parbox{0.92\linewidth}{\centering\vspace{1cm}Missing: images/struct\_col.png\vspace{1cm}}}%
} \\ \vspace{0.15cm}
\IfFileExists{images/all_struct_profile.png}{%
  \includegraphics[width=0.42\textwidth]{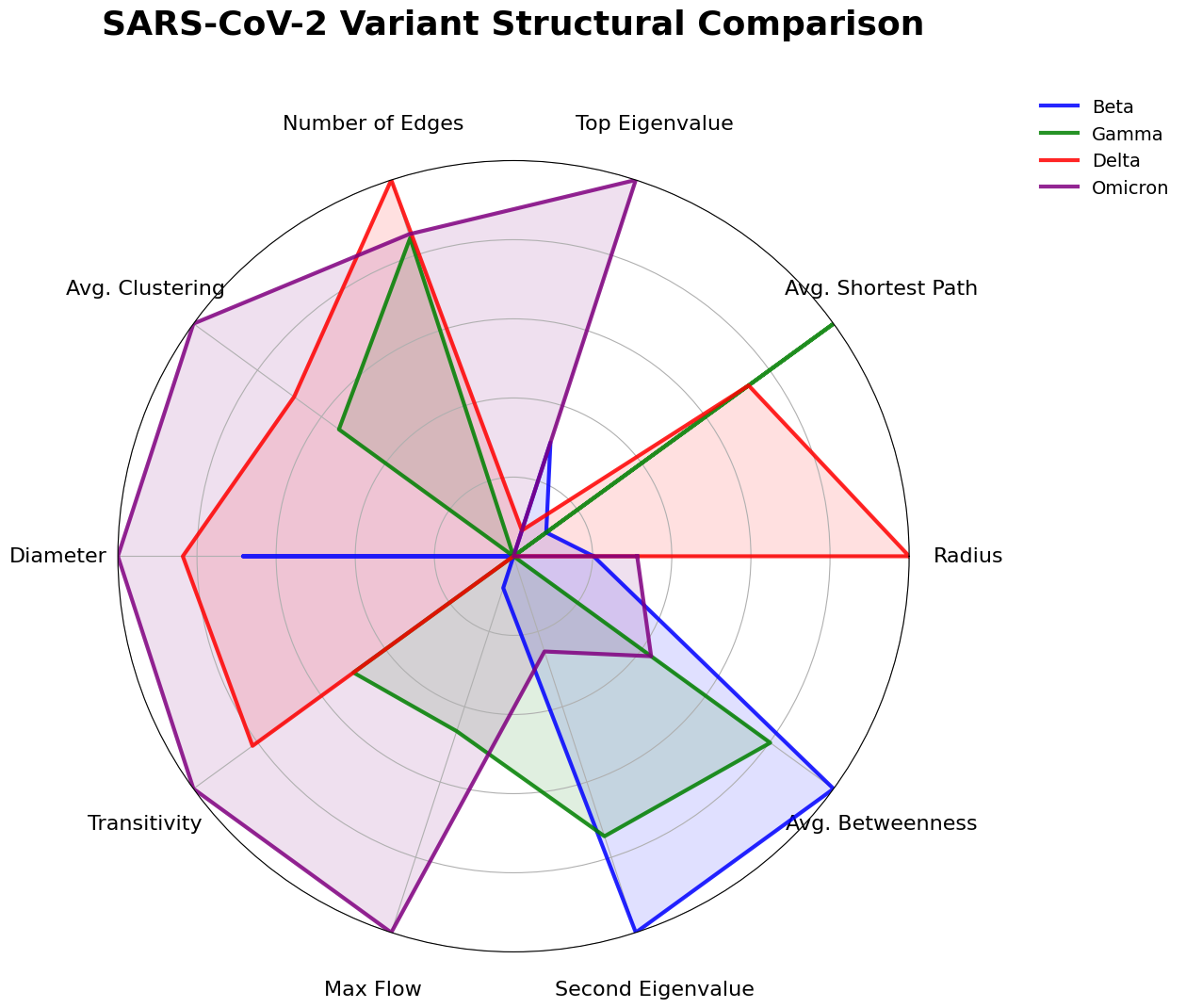}%
}{%
  \fbox{\parbox{0.5\linewidth}{\centering\vspace{1cm}Missing: images/all\_struct\_profile.png\vspace{1cm}}}%
}
\caption{Individual and Combined structural profiles for Beta, Gamma, Delta, and Omicron}
\label{fig:struct_col}
\end{figure*}

Figure~\ref{fig:struct_col} summarizes all four variant structural profiles in a single 2$\times$2 panel, so the visual comparison remains compact and the text flow stays uninterrupted.

\subsection{Cross-Variant Comparative Trend}

\begin{figure*}[!t]
\centering
\IfFileExists{images/paraller_cordinate_eviluton_graph.png}{%
  \includegraphics[width=\textwidth]{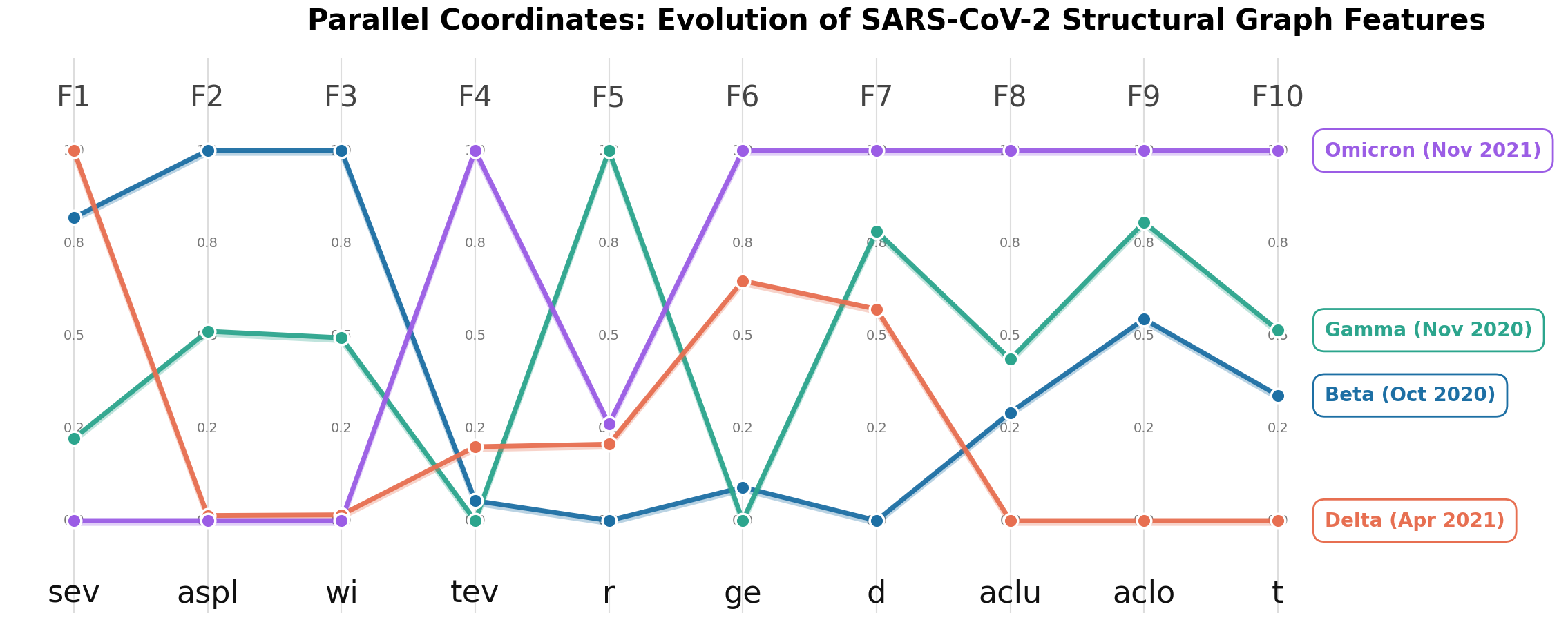}%
}{%
  \fbox{\parbox{0.95\linewidth}{\centering\vspace{1cm}Missing: images/paraller\_cordinate\_eviluton\_graph.png\vspace{1cm}}}%
}
\caption{Parallel-coordinate comparison of ten structural graph features across four SARS-CoV-2 variants of concern (Beta, Oct 2020; Gamma, Nov 2020; Delta, Apr 2021; Omicron, Nov 2021). Each polyline represents one variant; color encodes variant class. The ten axes correspond to: second eigenvalue (\textit{sev}), average shortest path length (\textit{aspl}), Wiener index (\textit{wi}), top eigenvalue (\textit{tev}), radius (\textit{r}), graph energy (\textit{ge}), diameter (\textit{d}), average clustering coefficient (\textit{aclu}), average closeness centrality (\textit{aclo}), and transitivity (\textit{t}). All values are min--max normalised to $[0,1]$. Omicron consistently occupies the highest band across spectral and centrality axes (\textit{ge}, \textit{aclu}, \textit{aclo}), while Delta remains near the minimum on most features, highlighting a sharp structural divergence between these two variants.}
\label{fig:parallel_evolution}
\end{figure*}

The parallel-coordinate view (Fig.~\ref{fig:parallel_evolution}) highlights how graph-derived features evolve differently across variants while preserving an overall shared structural backbone. Each polyline represents one genome sample; color encodes the variant class. This supports the idea that codon graph topology captures both conserved genome organization and variant-specific signatures simultaneously. Note how features such as matching number and stable rank sharply diverge for Omicron samples (visible in the upper and lower rows of the plot), while spectral measures such as top eigenvalue and graph energy remain highly conserved across all four clades.

\subsection{Interpretation}

\textbf{This section presents a biological and structural interpretation of the graph theoretic features extracted from the genome sequences of four SARS-CoV-2 variants: Beta (B.1.351), Gamma (P.1), Delta (B.1.617.2), and Omicron (B.1.1.529).\cite{ncbi}} Each genome sequence was represented as a graph, and 22 graphical properties were computed. The interpretation is organized into two subsections: (i) structural profile for each variant, (ii) comparative analysis of structural similarities and dissimilarities across the four variants.

\subsection{Structural Profiles of Individual Variants}

Each variant yields a distinct genomic graph fingerprint, defined by a combination of conserved baseline properties and variant-specific outlier behavior.

\subsubsection{Beta (B.1.351)}

\textbf{The Beta variant is characterized by a structurally stable median profile, combined with the most extreme individual outliers observed across the entire dataset, making it the most internally volatile variant at the graph level.}

From the graphs generated, it can be observed that the beta variants' genomic sequences produce dense interconnected networks with nearly complete graphs. This is reflected in the median graph density of $\sim 0.97$, which also indicates strong tendencies to conserve the viral backbone. Similarly, transitivity and average clustering coefficient values are about 0.96--0.97 across most Beta samples, confirming that local genomic neighborhoods are tightly triangulated. The average shortest path length and the median diameter of nearly 6.2 and 9.5, respectively, establish that the Beta genome graph exhibits a compact, small-world organization in which any genomic region is reachable from any other within a limited number of steps.

Regardless of the stable core, Beta sequences exhibit volatile-periphery anomalies, with several extreme outliers. From the graphical feature extraction, we have observed that at least one Beta sequence generates a Wiener Index of $\sim 45{,}000$, roughly thrice the typical value (vs. typical $\sim 15{,}000$) for other variants, indicating greater distances between genomic units. This longer genomic distance indicates extensive deletions or structural rearrangements, which elongated the internal pathways of the graph. Another similar phenomenon can be observed where one Beta sample produced a value of $\sim 4\times10^{10}$ (the highest in the dataset). Furthermore, a beta sequence dropped its average closeness centrality to 0.6, while the remaining clade clusters were near 0.98--0.99, indicating structurally remote regions of the graph from a connectivity-severing deletion. Finally, Beta's maximum flow distribution is the widest of all four variants (spanning about 9.2 to 10.0), showing highly variable capacity for parallel information propagation.

\subsubsection{Gamma (P.1)}

\textbf{Among the four variants, Gamma displays the broadest intra-clade structural diversity within its clade. While the outlying values for the Beta variant are found in only certain graphical features, Gamma distributes its structural variability across multiple graph properties simultaneously, producing the most heterogeneous collection of genome graphs in the dataset.}

Gamma has $\sim110$ nodes (an upward outlier), the highest in the dataset, compared to the average median of 65--70 across other variants, making Gamma a diverse and insertion-rich sample; the largest connected component of a Gamma sample reaches $\sim115$  nodes, the highest amongst all variants. This variant tends not to form isolated subgraphs; rather, it forms a single, unfragmented, interconnected giant graph. Furthermore, Gamma has the widest and highest diameter distribution among the variants, with outliers reaching up to 11.5. This physical elongation, along with the high betweenness centrality of Gamma, with outliers reaching up to $\sim 0.006$, implies that specific sites in the genomic sequences act as bridge notes controlling the information flow between distant genomic regions.

In addition, the graph energy of Gamma outliers reaches up to $\sim 175$, much higher than the typical 115--120 range. The higher structural complexity suggests a different eigenvalue distribution and indicates recombinant or heavily mutated genome sequences. Moreover, Gamma displays the lowest transitivity across the variants, outliers dropping to $\sim 0.84$, which suggests that local neighborhood structures are less triangulated and the otherwise conserved co-occurrence relationships have been disrupted. Overall, Gamma shows signs of extensive evolutionary development; the graphs span the widest range of organizational forms: elongated, bridge-dominated, and energy-rich structures. This structural variation depicts the different insertion events of the P.1 lineage, creating functional differences that require further investigation.

\subsubsection{Delta (B.1.617.2)}

\textbf{Delta is the most structurally disciplined variant in this dataset, producing genome graphs that are structurally consistent, compact, and well-clustered. }

Across almost every graph measure among the variants, Delta has the tightest distributions, intra-clade variance, and fewest extreme outliers, demonstrating the highest structural conservation. It yielded small and consistent networks with a median size of 65-70 nodes and approximately 2,000 edges, reflecting a marked degree of uniformity in structural magnitude. The distribution of its Wiener index is close to 15,000, indicating that the Delta genome graph is highly cohesive and replicable. Furthermore, Delta has a clustering coefficient with a median of $\sim 0.955$  (marginally lower than Beta and Gamma). The slight relaxation in triplet formation within this genomic network is consistent with the double-mutation feature of Delta (L452R and T478K), in which each region becomes slightly less mutually dependent on the others. Additionally, its spectral gap is the lowest among all four variants at $\sim 81.0$, indicating marginally less algebraic connectivity robustness.

In particular, Delta has no extreme outliers in either its condition number or its stable rank, which results  in yielding a consistently well-conditioned adjacency matrix with linearly independent, non-redundant, and non-degenerate structural motifs. Moreover, Delta also maintains an extremely low value for its matching number at $\sim 3$, which indicates that the genomic sequences of these variants' mutations do not produce independent modular components, keeping the genome a single, tightly integrated graph body. The sole exception is a single extreme-low-density outlier at $\sim 0.3$ (the typical value is $\sim 0.97$). This anomaly depicts a highly sparse graph with fewer edges; probably caused by sequencing artifact, highly defective genome, or large-scale deletion events rather than true mutation biology. Except for this outlier, Delta's profile can be defined by maximum conservation, achieving dominance by consolidating and optimizing a single, efficient genomic organization.

\subsubsection{Omicron (B.1.1.529)}

\textbf{The Omicron variant presents the most structurally complex and structurally modular genomic graph profile in this dataset.}

The most prominent graphical feature of Omicron is the matching number, which reaches up to $\sim 55$, whereas the other variants have a value of $\sim 3$, 18 times lower than Omicron. As the matching number indicates the size of an independent maximum edge set, this extreme escalation indicates that the genome graph contains a vast array of structurally non-overlapping, independent motifs. From a biological perspective, this is explained by the fact that Omicron has accumulated more than 30 spike protein mutations located at distinct independent sequence sites. This framework measures the independent evolutionary landscape of Omicron and depicts that its mutations are not clustered but are separate events.

Moreover, the structural diversification is evident in the increased stable rank, reaching $\sim 1.75$ for some Omicron sequences, whereas in other variants it lies in the 1.3--1.4 range. Stable rank indicates a higher effective dimensionality, which proves that Omicron graph structures occupy more complex and higher-dimensional spaces rather than being mere scale-ups of previous variants. This diversity is also shown by the increased Frobenius norm in Omicron graphs, which is $\sim 12.5$ (compared to the typical $\sim 10.5$), indicating greater structural mass and expansion in genomic connections. However, the modularity does not hinder global coherence as Omicron has the highest median average closeness centrality of $\sim 0.985$, ensuring that the nodes are well integrated and efficiently traversable. Omicron has the highest median spectral gap, along with Beta ($\sim 81.5$), and indicates the widest overall spread; this indicates a high algebraic strength, which will enable its functionality even when certain epitope-coding regions are disrupted. Overall, with transitivity and average clustering remaining around 0.96, except for one downward outlier, Omicron achieves a unique change. It maximizes structural modularity while keeping global resilience intact.

\subsection{Structural Similarities and Dissimilarities Across Variants}

After identifying the unique structural identities of the four variants individually, this section presents the comparative patterns that emerged across the variants. The comparative analysis has been categorized into four points: (i) conserved global structure, (ii) structural clustering through hierarchical analysis, (iii) patterns of divergence in specific variants, and (iv) the spectral characteristics unique to each variant.

\subsubsection{Conserved Global Topology Across Variants}

\textbf{One of the most prominent cross-variant observations is the presence of strong conservation, with a highly consistent global small-world topological architecture regardless of the strong evolutionary pressure.} For all the variants used in this study, Beta, Gamma, Delta, and Omicron, the genomic graphs demonstrate near-identical median values across fundamental graph theoretical features: median density of $\sim 0.95$--$0.97$, mean shortest path length of $\sim 6.2$, median diameter of $\sim 9.5$, and transitivity and average clustering coefficient $\sim 0.96$--$0.97$. Average betweenness centrality remains consistently low at around 0.001 across all variants, thus implying the absence of any dominant bottleneck in globally yet locally strongly connected graphs. This convergence of topological characteristics is supported by similar spectral clustering results, including the largest eigenvalue $\sim 85$--$86$, graph energy $\sim 115$--$120$, and consistency in the second eigenvalue around 5.3 across all four variants. This consistency of spectral graph features depicts that the structural complexity and graph energy remain unchanged even with the emergence of new mutations on the surface, thus indicating that the eigenvalue spectrum controls the functionality and dynamics of the genome, such as synchronization and diffusion within the SARS-CoV-2 genome.

\subsubsection{Hierarchical Clustering}

\textbf{Hierarchical clustering on the full set of graph-theoretic features produces a dendrogram that reveals two distinct structural clusters: Beta and Delta cluster together at a distance of approximately 5.7, while Gamma and Omicron cluster together at a distance of approximately 7.3. The two super-clusters merge at a distance of approximately 9.6, indicating a substantial structural divide between the two groups.} 

The Beta--Delta (distance $\sim 5.7$) pairing is notable for its disparity between phylogenetic classification based on nucleotide sequence data and clustering of structures, which classify Beta and Delta as the most structurally related pair, regardless of arising independently from separate regions and lineages. Thus, it can be said that Beta and Delta converged on a similar genomic graph through independent evolution; this phenomenon of convergence is probably driven by identical functional constraints and features: higher binding affinity with the ACE2 receptor and immunity evasions. This \{Beta, Delta\}  cluster represents an evolutionary strategy of structural consolidation, characterized by tighter distributions and fewer outlier events.

The Gamma---Omicron (distance $\sim 7.3$) pair is  also noteworthy as Gamma and Omicron share more structural similarity with each other than with Beta or Delta, despite Gamma emerging from the B.1.1.28 lineage in Brazil and Omicron emerging in South Africa and representing a deeply diverged lineage of uncertain ancestral origin. The grouping of these two strains could indicate convergence, common mutational patterns affecting genomically similar regions, or similar selective forces leading to structural convergence toward a similar set of structural features relevant to immune avoidance or transmission capabilities. The \{Gamma, Omicron\} cluster follows an evolutionary path of growing structural complexity and divergence, with greater structural diversity and outlier events.

Eventually, the two super-clusters {Beta, Delta} and {Gamma, Omicron} merge at $\sim 9.6$, representing a substantial structural discontinuity between the two.

\begin{figure}[!t]
\centering
\IfFileExists{images/Median.png}{%
  \includegraphics[width=\columnwidth]{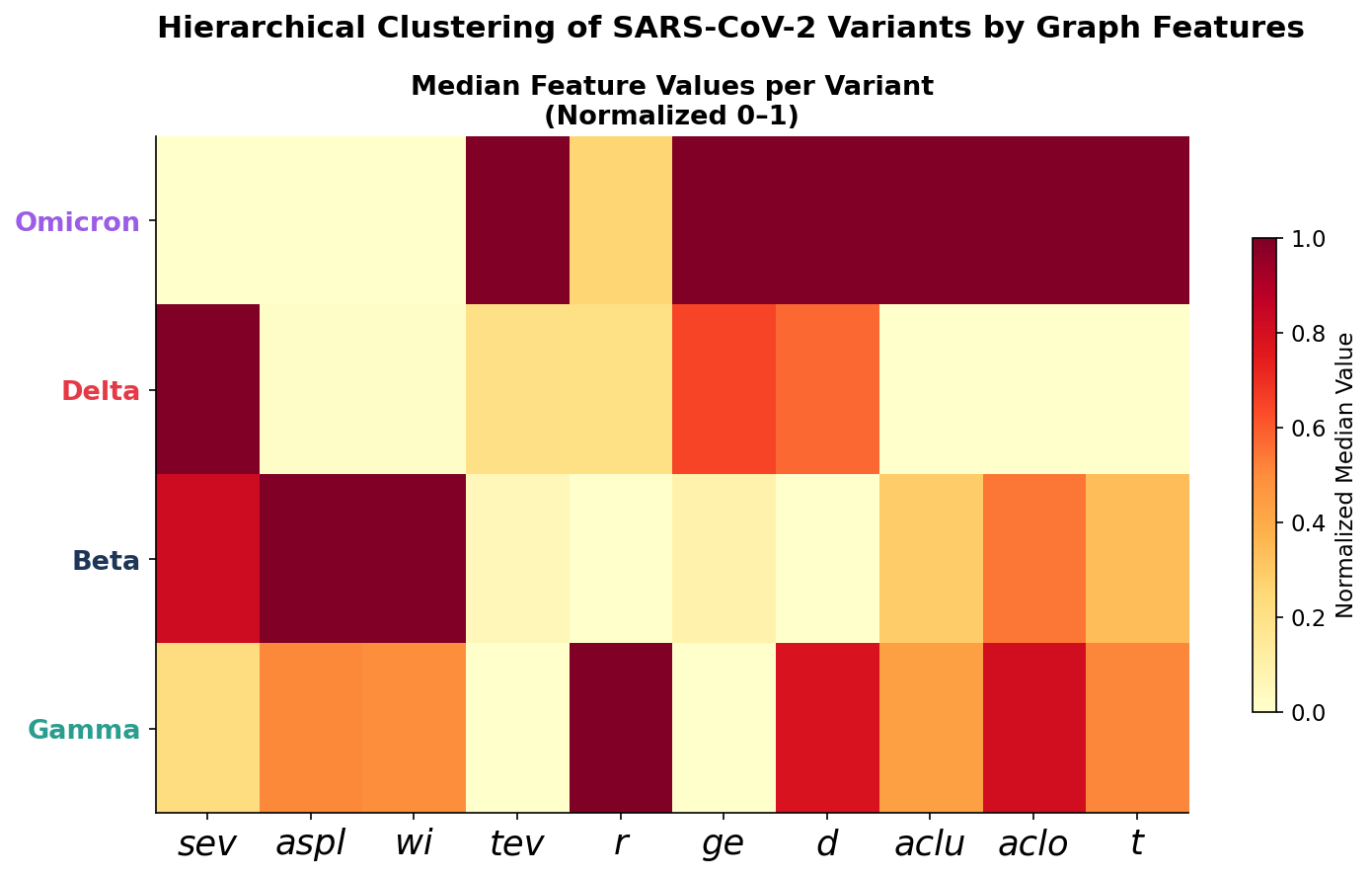}%
}{%
  \fbox{\parbox{0.95\linewidth}{\centering\vspace{1cm}Missing: images/Median.png}\vspace{1cm}}}%
\caption{Hierarchical Clustering of SARS-CoV-2 Variants by Graph Features. The ten features in x-axis correspond to: second eigenvalue (\textit{sev}), average shortest path length (\textit{aspl}), Wiener index (\textit{wi}), top eigenvalue (\textit{tev}), radius (\textit{r}), graph energy (\textit{ge}), diameter (\textit{d}), average clustering coefficient (\textit{aclu}), average closeness centrality (\textit{aclo}), and transitivity (\textit{t}). All values are min--max normalised to $[0,1]$. Omicron consistently occupies the highest band across spectral and centrality axes (\textit{ge}, \textit{aclu}, \textit{aclo}), while Delta remains near the minimum on most features, highlighting a sharp structural divergence between these two variants.}
\label{fig:Median}
\end{figure}

\subsubsection{Variant-Specific Structural Divergence}

\textbf{While the conserved global topology unifies all four variants at the median level, each variant diverges from the others in distinct ways that are specific to its graph-theoretic profile.} 

Beta diverges primarily through extreme outlier events in the Wiener index, condition number, and closeness centrality. These outliers occur independently and are not consistently co-occurring within the same samples, suggesting that Beta harbors multiple distinct mechanisms of structural disruption at the sublineage level. Beta's structural divergence is therefore characterized by intra-clade heterogeneity rather than a systematic shift of the entire clade away from the median.

Gamma diverges when multiple structural complexity features are simultaneously increased, including larger diameter, higher betweenness centrality, greater graph energy, and lower transitivity. These properties may all stem from a shared cause: the inclusion of genomic material. This elongates the graphs, creates bridge positions, raises the eigenvalue-based complexity, and disrupts local triangular structures. Thus, the divergence in Gamma is more structurally coherent than that of Beta: it shows a systematic structural reorganization in a subset of sequences rather than unrelated isolated changes.

Delta diverges minimally and specifically. Its only major cross-variant departure is the extreme low-density outlier, which is likely artifactual. In all other respects, Delta's distributions are the closest to the cross-variant consensus of any clade. Delta's structural divergence is negligible at the clade level, reinforcing its identity as the most evolutionarily conservative variant from a graph-theoretic perspective.

Omicron's matching number provides its clearest and most consistent divergence. Unlike the outlier events in Beta and Gamma, which are visible in only a fraction of samples, Omicron's elevated matching number is a systematic property that distinguishes the entire high end of the Omicron distribution from all other variants. Combined with its elevated stable rank, Omicron's divergence reflects a genuine structural transition---a qualitative change in the type of genome graph produced---rather than quantitative extremes within an otherwise familiar structural framework.

\subsubsection{Spectral and Algebraic Divergence}

\textbf{Spectral and algebraic analysis adds a second comparative layer.} Although the top eigenvalue is extremely conserved across all variants at a range of $\sim 85$--$86$, which indicates a strong global structural similarity, the spectral gap (measures algebraic connectivity robustness) differs significantly: both Beta and Omicron have the highest median values at about 81.5, Delta possesses the smallest at $\sim 81.0$, and Gamma falls between them. Condition number demonstrates the most divergent algebraic structure from the rest; while being relatively small and stable for all other variants, the outlying condition number of a near-singular Beta adjacency matrix approaches an astronomical value of roughly $4 \times 10^{10}$, suggesting a very brittle, algebraically sensitive graph prone to big perturbations. This creates a sharp contrast with the consistently well-conditioned Delta graphs and structurally robust Omicron graphs. Furthermore, while the Frobenius norm (reflects the overall magnitude of the adjacency matrix) remains relatively constant for all variants at about 10.5, Omicron displays its own set of outliers reaching 12.5. This elevated structural mass further corroborates Omicron's unique matching number and stable rank, reinforcing that it represents a quantitatively distinct, complex structural state at the algebraic level.

In summary, viral graph evolution appears to follow two concurrent trends: the conservation of structural backbone and variant-specific algebraic fingerprints. The clustering achieved by the GenEX pipeline differs from that of sequence-only phylogeny, indicating that graph-theoretic analysis captures an additional axis of viral evolution that complements sequence-based methods.

\section{Conclusion}
\label{sec:conclusion}
Here, we presented a unified benchmarking pipeline for the structural analysis of SARS-CoV-2 variants: Beta, Delta, Gamma, and Omicron, at the nucleotide and codon level, through graphical analysis of the codon interaction network system. We introduced two novel algorithms to overcome the limitations of linear sequence analysis, currently used in genome  sequence analysis: \textbf{(i) Linear-time Adjacency PMI Codon Graph (LAPCG)} and \textbf{(ii) Multi-Scale Codon Co-occurrence Graph (MSCG)}. By representing genomic sequences as interaction networks and extracting spectral and topological features, we demonstrated that variants possess unique \textbf{structural fingerprints} that can be accurately classified using machine learning models such as Random Forests, Extra Trees, and LightGBM. The structural interpretation of these networks provided critical insights into the virus's evolutionary trajectory.

\textbf{LAPCG} is a well-optimized algorithm for graph extraction and feature computation that uses direct-neighbor approximation to achieve linear-time complexity O(n). In contrast, \textbf{MSCG} is a mathematically principled approach with the capability to model complex inter-context dependencies at multiple co-occurrence scales. Based on our results, \textbf{MSCG} (utilizing Normalized PMI and positional node encoding) has maintained an accuracy of \textbf{98.75\%}, hence establishing a robust, alignment-free framework for genomic analysis. Although currently focused on codon transition patterns, this work identifies vital topological biomarkers and explainability of viral genome sequences across clades and generations, using mathematical and graphical models. Future research will focus on integrating protein structural data and time-aware viral genomic evolution, and on extending this pipeline to other viruses and higher organisms.

\bibliographystyle{ieeetr}
\bibliography{conf}

@article{zhou2015predicting,
  author  = {Zhou, J. and Troyanskaya, O.},
  title   = {Predicting effects of noncoding variants with deep learning–based sequence model},
  journal = {Nature Methods},
  volume  = {12},
  pages   = {931--934},
  year    = {2015},
  doi     = {10.1038/nmeth.3547},
  url     = {https://doi.org/10.1038/nmeth.3547}
}

@article{avsec2021effective,
  author  = {Avsec, Ž. and Agarwal, V. and Visentin, D. and others},
  title   = {Effective gene expression prediction from sequence by integrating long-range interactions},
  journal = {Nature Methods},
  volume  = {18},
  pages   = {1196--1203},
  year    = {2021},
  doi     = {10.1038/s41592-021-01252-x},
  url     = {https://doi.org/10.1038/s41592-021-01252-x}
}

@article{garrison2018variation,
  author  = {Garrison, E. and Sir{\'e}n, J. and Novak, A. M. and Hickey, G. and Eizenga, J. M. and Dawson, E. T. and Jones, W. and Garg, S. and Markello, C. and Lin, M. F. and Paten, B. and Durbin, R.},
  title   = {Variation graph toolkit improves read mapping by representing genetic variation in the reference},
  journal = {Nature Biotechnology},
  volume  = {36},
  number  = {9},
  pages   = {875--879},
  year    = {2018},
  month   = oct,
  doi     = {10.1038/nbt.4227},
  pmid    = {30125266},
  pmcid   = {PMC6126949}
}

@article{zitnik2018modeling,
  title     = {Modeling polypharmacy side effects with graph convolutional networks},
  volume    = {34},
  ISSN      = {1367-4811},
  url       = {http://dx.doi.org/10.1093/bioinformatics/bty294},
  DOI       = {10.1093/bioinformatics/bty294},
  number    = {13},
  journal   = {Bioinformatics},
  publisher = {Oxford University Press (OUP)},
  author    = {Zitnik, Marinka and Agrawal, Monica and Leskovec, Jure},
  year      = {2018},
  month     = jun,
  pages     = {i457--i466}
}

@inproceedings{gilmer2017neural,
  author    = {Gilmer, Justin and Schütt, Kristof T. and Glawe, Patrick and Klambauer, Günter and Smola, Alexander and Welling, Max},
  title     = {Neural Message Passing for Quantum Chemistry},
  booktitle = {Proceedings of the 34th International Conference on Machine Learning},
  series    = {ICML},
  year      = {2017},
  url       = {https://arxiv.org/abs/1704.01212}
}

@incollection{dayhoff1978atlas,
  author    = {Dayhoff, M. O. and Schwartz, R. M. and Orcutt, B. C.},
  title     = {A model of evolutionary change in proteins},
  booktitle = {Atlas of Protein Sequence and Structure},
  editor    = {Dayhoff, M. O.},
  publisher = {National Biomedical Research Foundation},
  address   = {Washington DC},
  volume    = {5},
  number    = {3},
  pages     = {345--352},
  year      = {1978}
}

@article{henikoff1992amino,
  author  = {Henikoff, S. and Henikoff, J. G.},
  title   = {Amino acid substitution matrices from protein blocks},
  journal = {Proceedings of the National Academy of Sciences of the United States of America},
  volume  = {89},
  number  = {22},
  pages   = {10915--10919},
  year    = {1992},
  month   = nov,
  doi     = {10.1073/pnas.89.22.10915},
  pmid    = {1438297},
  pmcid   = {PMC50453}
}

@misc{shrikumar2017learning,
  title         = {Learning Important Features Through Propagating Activation Differences},
  author        = {Avanti Shrikumar and Peyton Greenside and Anshul Kundaje},
  year          = {2019},
  eprint        = {1704.02685},
  archivePrefix = {arXiv},
  primaryClass  = {cs.CV},
  url           = {https://arxiv.org/abs/1704.02685}
}

@article{burge1997prediction,
  author  = {Burge, C. and Karlin, S.},
  title   = {Prediction of complete gene structures in human genomic DNA},
  journal = {Journal of Molecular Biology},
  volume  = {268},
  number  = {1},
  pages   = {78--94},
  year    = {1997},
  month   = apr,
  doi     = {10.1006/jmbi.1997.0951},
  pmid    = {9149143}
}

@article{lomsadze2005gene,
  author  = {Lomsadze, A. and Ter-Hovhannisyan, V. and Chernoff, Y. O. and Borodovsky, M.},
  title   = {Gene identification in novel eukaryotic genomes by self-training algorithm},
  journal = {Nucleic Acids Research},
  volume  = {33},
  number  = {20},
  pages   = {6494--6506},
  year    = {2005},
  month   = nov,
  doi     = {10.1093/nar/gki937},
  pmid    = {16314312},
  pmcid   = {PMC1298918}
}

@article{korf2004gene,
  author  = {Korf, I.},
  title   = {Gene finding in novel genomes},
  journal = {BMC Bioinformatics},
  volume  = {5},
  pages   = {59},
  year    = {2004},
  month   = may,
  doi     = {10.1186/1471-2105-5-59},
  pmid    = {15144565},
  pmcid   = {PMC421630}
}

@article{majoros2004tigrscan,
  author  = {Majoros, W. H. and Pertea, M. and Salzberg, S. L.},
  title   = {{TigrScan} and {GlimmerHMM}: two open source ab initio eukaryotic gene-finders},
  journal = {Bioinformatics},
  volume  = {20},
  number  = {16},
  pages   = {2878--2879},
  year    = {2004},
  month   = nov,
  doi     = {10.1093/bioinformatics/bth315},
  pmid    = {15145805}
}

@article{holt2011maker,
  author  = {Holt, C. and Yandell, M.},
  title   = {{MAKER2}: an annotation pipeline and genome-database management tool for second-generation genome projects},
  journal = {BMC Bioinformatics},
  volume  = {12},
  pages   = {491},
  year    = {2011},
  doi     = {10.1186/1471-2105-12-491},
  url     = {https://doi.org/10.1186/1471-2105-12-491}
}

@article{stanke2008using,
  title   = {Using native and syntenically mapped {cDNA} alignments to improve de novo gene finding},
  author  = {Mario Stanke and Mark E. Diekhans and Robert Baertsch and David Haussler},
  journal = {Bioinformatics},
  year    = {2008},
  volume  = {24},
  number  = {5},
  pages   = {637--644},
  url     = {https://api.semanticscholar.org/CorpusID:16520710}
}

@incollection{hoff2019whole,
  author    = {Hoff, K. J. and Lomsadze, A. and Borodovsky, M. and Stanke, M.},
  title     = {Whole-Genome Annotation with {BRAKER}},
  booktitle = {Methods in Molecular Biology},
  volume    = {1962},
  pages     = {65--95},
  year      = {2019},
  doi       = {10.1007/978-1-4939-9173-0_5},
  pmid      = {31020555},
  pmcid     = {PMC6635606}
}

@article{holst2023helixer,
  author   = {Stiehler, Felix and Steinborn, Marvin and Scholz, Stephan and Dey, Daniela and Weber, Andreas P M and Denton, Alisandra K},
  title    = {Helixer: cross-species gene annotation of large eukaryotic genomes using deep learning},
  journal  = {Bioinformatics},
  volume   = {36},
  number   = {22-23},
  pages    = {5291--5298},
  year     = {2021},
  month    = apr,
  doi      = {10.1093/bioinformatics/btaa1044},
  url      = {https://doi.org/10.1093/bioinformatics/btaa1044},
  issn     = {1367-4803}
}

@misc{ncbi,
  author       = {{National Center for Biotechnology Information}},
  title        = {NCBI},
  year         = {2026},
  howpublished = {\url{https://www.ncbi.nlm.nih.gov/}},
  note         = {Accessed: 2026-02-22}
}

\end{document}